\documentclass{article}

\usepackage{microtype}
\usepackage{graphicx}
\usepackage{subfigure}
\usepackage{booktabs} 

\usepackage{hyperref}

\usepackage[accepted]{icml2023}

\usepackage{amsmath}
\usepackage{amssymb}
\usepackage{mathtools}
\usepackage{amsthm}

\usepackage[capitalize,noabbrev]{cleveref}
\PassOptionsToPackage{sort&compress}{natbib}

\theoremstyle{plain}

\theoremstyle{definition}

\theoremstyle{remark}

\usepackage[textsize=tiny]{todonotes}

\icmltitlerunning{Tune As You Scale: Hyperparameter Optimization For Compute Efficient Training}

\begin{document}

\twocolumn[
\icmltitle{Tune As You Scale: \\ Hyperparameter Optimization For Compute Efficient Training}



\icmlsetsymbol{equal}{*}

\begin{icmlauthorlist}
\icmlauthor{Abraham J. Fetterman}{Generally Intelligent}
\icmlauthor{Ellie Kitanidis}{Generally Intelligent}
\icmlauthor{Joshua Albrecht}{Generally Intelligent}
\icmlauthor{Zachary Polizzi}{Generally Intelligent}
\icmlauthor{Bryden Fogelman}{Generally Intelligent}
\icmlauthor{Maksis Knutins}{Generally Intelligent}
\icmlauthor{Bartosz Wróblewski}{Generally Intelligent}
\icmlauthor{James B. Simon}{Generally Intelligent,UC Berkeley}
\icmlauthor{Kanjun Qiu}{Generally Intelligent}
\end{icmlauthorlist}

\icmlaffiliation{Generally Intelligent}{Generally Intelligent}
\icmlaffiliation{UC Berkeley}{UC Berkeley}

\icmlcorrespondingauthor{Abraham J. Fetterman}{abe@generallyintelligent.com}

\icmlkeywords{Machine Learning, ICML}

\vskip 0.3in
]



\printAffiliationsAndNotice{}  

\begin{abstract}
Hyperparameter tuning of deep learning models can lead to order-of-magnitude performance gains for the same amount of compute. Despite this, systematic tuning is uncommon, particularly for large models, which are expensive to evaluate and tend to have many hyperparameters, necessitating difficult judgment calls about tradeoffs, budgets, and search bounds. To address these issues and propose a practical method for robustly tuning large models, we present \textbf{C}ost-\textbf{A}ware Pareto \textbf{R}egion \textbf{B}ayesian \textbf{S}earch (CARBS), a Bayesian optimization algorithm that performs local search around the performance-cost Pareto frontier. CARBS does well even in unbounded search spaces with many hyperparameters, learns scaling relationships so that it can tune models even as they are scaled up, and automates much of the ``black magic'' of tuning. Among our results, we effectively solve the entire ProcGen benchmark just by tuning a simple baseline (PPO, as provided in the original ProcGen paper). We also reproduce the model size vs. training tokens scaling result from the Chinchilla project \cite{Hoffmann++22}, while simultaneously discovering scaling laws for every other hyperparameter, via an easy automated process that uses significantly less compute and is applicable to any deep learning problem (not just language models).
\end{abstract}

\section{Introduction}
\label{sec:intro}

Deep learning requires hyperparameter tuning, which is a black-box optimization problem with expensive evaluations. Tuning simple baselines can lead to significantly better performance for the same amount of compute \cite{ResNet-RS, Hoffmann++22}. A striking recent demonstration of this is the scaling laws study from DeepMind's Chinchilla project \cite{Hoffmann++22}, which showed that a 70B parameter language model can outperform a 175B model when the number of training tokens is properly tuned. Though several approaches for tuning exist, Bayesian optimization (BO) using Gaussian process (GP) surrogate models (see e.g. \citealt{Shahriari++16} for a review) has emerged as the primary tool thanks to its sample efficiency, flexibility, and robustness \cite{Snoek++12, Turner++21}. 

However, despite the proven value and an ever-expanding zoo of tuning algorithms, networks today are still often not exhaustively tuned. In an informal survey of researchers presented at NeurIPS 2022 \cite{Schneider++22}, approximately three-quarters of respondents reported that they tune five or fewer hyperparameters, and nearly half stated that they perform tuning manually. According to the survey, fewer than a third of researchers use more than 25 tuning trials. Aside from potentially missing out on important discoveries like the Chinchilla scaling laws, a lack of thorough tuning is concerning because it can lead to suboptimal performance, unnecessary expense, and ambiguity when comparing new methods to the previous state-of-the-art. 

There are several practical limitations to current hyperparameter tuning approaches. First, as deep learning models are scaled up, evaluations become prohibitively expensive. The standard practice is to tune smaller models and then naively extrapolate to larger models, or to use the hyperparameters of the smaller models as a ``warm start'' and do a minimal amount of additional tuning for larger models. However, there is considerable evidence that the optimal hyperparameters depend on scale \cite{Kaplan++20, Hoffmann++22, Yang++22}. In order to efficiently tune large models, the tuner must learn these scaling relationships, exploiting statistical information gained from iterating on smaller models.

Additionally, although BO has been very successful on problems with a small number of parameters, it often performs poorly as the number of parameters grows (see e.g. \citealt{Wang+13}), in part because the search space grows exponentially. Global acquisition functions tend to over-explore larger search spaces since there are more highly uncertain regions. For this reason, often only a small subset of hyperparameters are tuned in practice.

These constraints inhibit fully-automated hyperparameter tuning, as decisions about which subsets of hyperparameters to tune, how to balance performance against cost, and what scaling strategies to employ require considerable manual effort and expertise. In addition to being time-consuming, manual tuning is fundamentally a bias-prone and irreproducible way of doing science.

To overcome these limitations, we present \textbf{C}ost-\textbf{A}ware Pareto \textbf{R}egion \textbf{B}ayesian \textbf{S}earch (CARBS). CARBS is a BO algorithm that models both performance and cost (measured in training time) as GPs and jointly optimizes them. The acquisition function uses local search in the neighborhood of the current performance-cost Pareto frontier. This local search helps it optimize even in high-dimensional spaces. Furthermore, it is able to tune as it scales and can even predict scaling laws for every hyperparameter from the tuning process because it explicitly learns cost-dependent relationships. As such, it is entirely automated and does not require specifying search bounds or model training budgets. This also allows it to tune parameters such as the number of training epochs or tokens, which are usually held constant in the tuning process even though, as \citealt{Hoffmann++22} proved, co-tuning these hyperparameters can dramatically improve both the performance and compute efficiency of models.

We empirically demonstrate that on challenging open problems such as ProcGen \cite{Cobbe++19}, running CARBS on one of the simplest baselines, Proximal Policy Optimization (PPO; \citealt{Schulman++17}), leads to effectively solving the entire benchmark. Furthermore, we reproduce the scaling laws of \citealt{Hoffmann++22} by simply running CARBS to tune GPT \cite{GPT} on C4 \cite{C4}. Even on well-tuned smaller problems, CARBS achieves similar performance significantly faster and more consistently. 

\section{Related Work}
\label{sec:related}

\textbf{Local search.} The notion of local stochastic search with an iteratively updated search distribution is common in evolutionary strategies such as Covariance Matrix Adaptation (CMA-ES; \citealt{CMAES}), though such methods typically do not learn a surrogate model of the objective function. A few variants of BO with local search also exist, e.g. using a collection of local models in rectangular trust regions \cite{Turbo} or restricting to samples the GP is certain about \cite{Fröhlich++21}. To the best of our knowledge, our work appears to be novel in its application of local search to the Pareto front of past observations.

\textbf{Cost-aware BO.} The works most related to ours leverage some degree of cost-awareness in the search strategy. \citealt{Abdo20} explicitly searches for cheap solutions first and requires users to input an ordering of the dimensions of the search space based on prior knowledge of their relative expense. Several works \cite{Snoek++12, Swersky++13, Poloczek++16, Wu++19, Lee++20} use a canonical acquisition function such as the Expected Improvement (EI) divided by a heterogeneous cost metric. However, these methods often show poor overall performance when the best hyperparameters are also the most expensive, as they tend to oversample cheap candidates. To reduce this effect, \citealt{Lee++20} implements a cost-cooling trick that transitions from cost-normalized EI to plain EI, though this requires knowing the compute budget a priori. \citealt{Guinet++20} minimizes cost constrained by EI, using a subset of candidates with sufficiently high EI at each step. \citealt{Lee++21} and \citealt{Astudillo+21} formulate the problem as a Markov decision process where a ``non-myopic'' acquisition function uses rollouts from the surrogate. These methods focus on finding the cheapest solutions that still perform well or the best performing methods within a given compute budget rather than on learning the structure of the Pareto front itself.

\textbf{Neural scaling laws.} A body of recent work focuses on the empirical scaling relationships between a network's performance, size, compute budget, and the amount of training data it sees. Influential early works \cite{Hestness++17, Rosenfeld++19, Tan++19, TrainLarge, Kaplan++20} laid the groundwork and a number of subsequent works have explored scaling laws across several modalities, architectures, and problem settings (e.g. \citealt{Tay++21, Zhai++21, Henighan++20, Droppo++21, Cherti++22}). A notable reminder of the importance of accurate neural scaling laws is \citealt{Hoffmann++22}, which demonstrated superior performance to its compute-equivalent precursor Gopher with a model that was 4x smaller. Scaling studies typically involve training large families of models across wide ranges of dataset sizes and compute budgets, and then analytically fitting the results. To draw reliable conclusions, each evaluation should ideally be well-tuned in the other hyperparameters, further adding to the computational burden. In practice, most studies are restricted to systematically varying a few dimensions at a time while holding the others fixed, which can miss improvements from correlated changes to several variables.

\section{Algorithm}
\label{sec:algo}

CARBS is a cost-aware optimization algorithm that is local to the compute-efficient Pareto front. We first give a simplified overview of the algorithm, then add some finer details in the following sections.

\subsection{Simplified algorithm}

We are interested in maximizing a function $f$ of real input parameters $\mathbf{x}$ that we observe through a noisy channel $y=f\left(\mathbf{x}\right)+\epsilon_{y}$ where $\epsilon_{y}$ is a mean zero Gaussian. In addition to the output of the function, we observe the cost $c$ of evaluating the function. We assume without loss of generality that our input parameters and outputs $\mathbf{x}$ and $y$ are scaled to be of order one.

The role of this algorithm is to suggest parameters $\mathbf{x}^{\prime}$ that will maximize $f(\mathbf{x}^{\prime})$, ignoring, for now, other heuristics related to the cost of evaluation. At a high level, CARBS works by first generating candidates from a local search space, evaluating those candidates with GP surrogates, then selecting the highest scoring candidate as scored by an acquisition function based on those surrogates. Each of these steps is explained more precisely below.

\textbf{Generating candidates in the local search space}

CARBS defines the local search space around points on the observed cost-performance Pareto front. If we assume we have $t$ observations, and that observation $i$ has parameters $\mathbf{x}_i$, output $y_i$ and cost $c_i$, then the Pareto front is

\begin{equation}
\label{eq:pf}
\text{PF} = \{ i \in [1:t] \ \text{s.t.} \ \{ y_{i}>y_{j}\,\lor\,c_{i}<c_{j} \} \ \forall j \neq i \}.
\end{equation}

We define the search space as a set of Gaussian distributions of radius $\sigma_{\mathrm{search}}$ around these parameters, defining the unnormalized probability density of $\mathbf{x}$ as

\begin{equation}
P_{\mathrm{search}}\left(\mathbf{x}\right)=\max_{i\in \rm{PF}}\left[\mathrm{exp}\left(-\frac{\left|\mathbf{x}_{i}-\mathbf{x}\right|^2}{2\sigma_{\mathrm{search}}^2}\right)\right].
\end{equation}

Candidates for the next suggestion are sampled uniformly from the Gaussian distributions  $\mathcal{N}\left(\mathbf{x}_{i},\sigma_{\mathrm{search}}\right)$. From an evolutionary strategies perspective, these can be described as mutations from a dynamic pool of parents (the Pareto set).

\textbf{Evaluating candidates via Gaussian process surrogates}

Once we have the candidate suggestions, we use GP surrogates to estimate the cost $\tilde{c}$ and output $\tilde{y}$ based on the parameters $\mathbf{x}$ of each candidate. CARBS uses three GPs: $\mathcal{GP}_{y}$ to predict the performance of a candidate, informed by all observations so far; $\mathcal{GP}_{c}$ to predict the cost of a candidate, informed by all observations so far; and $\mathcal{GP}_{\mathrm{pf}}$ to predict the optimal performance corresponding to that cost, informed only by the Pareto front so far. This last GP can be thought of as modeling the shape of the Pareto front and is used to calculate the baseline against which performance is compared in the EI acquisition function. 

For these surrogates, we utilize the same kernel function used in Heteroskedastic Evolutionary Bayesian Optimization (HEBO; \citealt{Cowen-Rivers++20}), the sum of a linear and Matern kernel. We find that the inclusion of the linear kernel in particular is crucial for getting good performance, as without it the model extrapolates poorly. The third GP, which estimates the Pareto front based only on the cost, uses an RBF kernel. In summary, we fit the GP models using the Gaussian process likelihood $p$ and kernels $k_\theta$,

\begin{equation}
\label{eq:gps}
\begin{aligned}
&\mathcal{GP}_{y} \ \leftarrow\max_{\theta}\left[p\left(y|\{\mathbf{x}_{i},y_{i}\}_{i\in\left[1:t\right]},k_{\theta}=k_{\mathrm{lin}}+k_{\mathrm{Mat}}\right)\right],\\
&\mathcal{GP}_{c} \   \leftarrow\max_{\theta}\left[p\left(c|\{\mathbf{x}_{i},c_{i}\}_{i\in\left[1:t\right]},k_{\theta}=k_{\mathrm{lin}}+k_{\mathrm{Mat}}\right)\right],\\
&\mathcal{GP}_{\mathrm{pf}} \leftarrow\max_{\theta}\left[p\left(y|\{c_{i},y_{i}\}_{i\in PF},k_{\theta}=k_{\mathrm{RBF}}\right)\right].
\end{aligned}
\end{equation}

In the following notation, we use $\mathcal{GP}_y(\mathbf{x})$ to indicate the posterior distribution of $\mathcal{GP}_y$ evaluated at $\mathbf{x}$.

\textbf{Scoring candidates via the acquisition function}

The final model $\mathcal{GP}_{\mathrm{pf}}$ is fit only using observations that belong in the Pareto front, and is used as the baseline for the expected improvement (EI) in the acquisition function in composition with the cost model. Thus, the acquisition function explicitly seeks to improve performance relative to the corresponding point on the Pareto front. That is, with the expected cost $\tilde{c}\left(\mathbf{x}\right)=\mathop{\mathbb{E}}\left[\mathcal{GP}_{c}\left(\mathbf{x}\right)\right]$ we can use the Pareto value  $\tilde{y}_{\mathrm{pf}}\left(\mathbf{x}\right)=\mathop{\mathbb{E}}\left[\mathcal{GP}_{\mathrm{pf}}\left(\tilde{c}\left(\mathbf{x}\right)\right)\right]$ to write the expected improvement with respect to the Pareto front,

\begin{equation}
\label{eq:ei-pf}
\alpha_{\mathrm{EI-pf}}\left(\mathbf{x}\right)=\mathop{\mathbb{E}}_{\mathcal{GP}_{y}}\left[\mathrm{ReLU}\left(\mathcal{GP}_{y}\left(\mathbf{x}\right)-\tilde{y}_{\mathrm{pf}}\left(\mathbf{x}\right)\right)\right],    
\end{equation}

where ReLU is the rectified linear unit and the outer expectation is over the distribution output by $\mathcal{GP}_{y}$. We combine this acquisition function with the search space probability naively by taking the product, and the final suggestion is simply the max over all candidates:

\begin{equation}
\alpha_{\mathrm{pf}}\left(\mathbf{x}\right)=\alpha_{\mathrm{EI-pf}}\left(\mathbf{x}\right)P_{\mathrm{search}}\left(\mathbf{x}\right).
\end{equation}

This effectively adds a soft trust region to the EI acquisition function. A hard trust region would discard any solution greater than some distance from a PF point, which could be implemented by using a uniform probability distribution within a sphere of radius $\sigma$ for $P_{\rm search}$. Instead of a hard cutoff, we discount points further from the trust region by using a Gaussian probability distribution.

The above acquisition function is enough to define a simplified version of CARBS found in the Appendix as Algorithm~\ref{alg:carbs-simple}. The next section fills in details that improve performance on practical problems.

\subsection{Additional Details}

\textbf{Acquisition function clamping.} An issue with the acquisition function  $\alpha_{\mathrm{EI-pf}}$ is that it uniformly weights improvements across the entire Pareto front, while we are more interested in the best performing parameters. The regular expected improvement uses $y_{\mathrm{max}}=\max\left(y_{i}\right)$,

\begin{equation}
\alpha_{\mathrm{EI-max}}\left(\mathbf{x}\right)=\mathop{\mathbb{E}}_{\mathcal{GP}_{y}}\left[\mathrm{ReLU}\left(\mathcal{GP}_{y}\left(\mathbf{x}\right)-y_{\mathrm{max}}\right)\right].
\end{equation}

We seek a balance between these two functions: empirically, $\alpha_{\mathrm{EI-max}}$ only samples near the highest performing part of the Pareto front, and  $\alpha_{\mathrm{EI-pf}}$ only rarely samples there. Our solution is to sample a threshold cost $c_{\mathrm{th}}\sim\mathrm{LogUniform}\left(\min_{PF}\left(c_{i}\right),\max_{PF}\left(c_{i}\right)\right)$ from a log uniform distribution from the minimum Pareto front cost to the maximum Pareto front cost. We then evaluate the Pareto front performance there, $\tilde{y}_{\mathrm{th}}=\mathop{\mathbb{E}}\left[\mathcal{GP}_{\mathrm{pf}}\left(c_{\mathrm{th}}\right)\right]$,

\begin{equation}
\alpha_{\mathrm{EI-th}}\left(\mathbf{x}\right)=\mathop{\mathbb{E}}_{\mathcal{GP}_{y}}\left[\mathrm{ReLU}\left(\mathcal{GP}_{y}\left(\mathbf{x}\right)-\max\left(\tilde{y}_{\mathrm{pf}}\left(\mathbf{x}\right),\tilde{y}_{\mathrm{th}}\right)\right)\right].
\end{equation}

See experiments comparing  $\alpha_{\mathrm{EI-th}}$ with  $\alpha_{\mathrm{EI}}$ and $\alpha_{\mathrm{EI-pf}}$ in Section \ref{subsec:exp-comparison}. In all other experiments, $\alpha_{\mathrm{EI-th}}$ is used.

\textbf{Output warping.} The GP models depend on the output variable having an approximately Gaussian distribution with zero mean and unit variance. HEBO showed that using a Box-Cox or Yeo-Johnston transform could improve performance of their Bayesian optimization algorithm~\cite{Cowen-Rivers++20}. We use a similar, parameter free method, the \texttt{scikit-learn}~\cite{scikit-learn} \texttt{QuantileTransform}\footnote{QuantileTransform achieves the same goal as the PowerTransform used by HEBO but is more numerically stable.} with $\sqrt{t}$ bins, to transform the output value $y$. We find the use of the parameter free transform to be more stable and repeatable than the learned parameter transforms.

\textbf{Resampling.} Noisy observations are particularly important to handle correctly for CARBS, as the best observations are used to define the local search space and thus a particularly extreme noise point might prevent the algorithm from finding a better optimum. To mitigate this issue, points on the Pareto front are resampled, or suggested again, at a fixed rate: one out of every $N_{\mathrm{resample}}$ suggestions is a set of parameters that has already been observed. Specifically, we count the samples for each of the existing Pareto front points and suggest the parameters which have the smallest number of samples, breaking ties by choosing the one with the lowest cost.

This resampling also allows us to more carefully select our Pareto set. We first group all observations by parameters, and then calculate the mean output $\bar{y}$, max output $y^\mathrm{+}$, average cost $\bar{c}$ and number of samples $n$ for each group. We then define the grouped Pareto set as

\begin{equation}
\begin{aligned}
PF_{\mathrm{group}}=
\{ 
i&\in[1:g]\,\mathrm{s.t.} \\
    &\begin{cases}   y_{i}^{\mathrm{+}}>y_{j}^{\mathrm{+}}\,\lor\,\bar{c}_{i}<\bar{c}_{j} \hspace{0.1cm} &(n_{i}=1)\\ \bar{y}_{i}>\bar{y}_{j}\,\lor\,\bar{c}_{i}<\bar{c}_{j} \hspace{0.1cm} &(n_{i}>1)
\end{cases} \\ 
&\hspace{0.2cm} \forall j\neq i \}.
\end{aligned}
\end{equation}

This includes single observations in the Pareto front only if they are better than the best observation that is lower cost, but allows resampled groups if their mean is higher than the best group mean that is lower cost.

\textbf{Pareto front minimum.} Because the cost itself is a noisy parameter, the very lowest cost observation is usually not very interesting to include in our Pareto front. We start the Pareto front with best performing observation among the 20\% of observations with the lowest cost.

\textbf{Failure prediction.} Tuning parameters like model size can be difficult if some parameters lead to failure (e.g., from out-of-memory errors). Including these as observations with poor results (for example, setting $y_i=0$) results in GP models that are not accurate near failure boundaries. We instead train a separate GP regression model to predict whether training will fail, setting $f_i=1$ if the result of a run is a failure or $f_i=-1$ if it is a success,

\begin{equation}
\begin{aligned}
\mathcal{GP}_{\mathrm{fail}} \leftarrow \max_{\theta} [ p (f | \ \{\mathbf{x}_{i},f_{i}\}_{i\in[1:t]},
k_{\theta}=k_{\mathrm{lin}}+k_{\mathrm{Mat}} )].
\end{aligned}
\end{equation}

We then calculate the probability of success as

\begin{equation}    
P_{\mathrm{success}}\left(\mathbf{x}\right)=\mathop{\mathbb{E}}_{\mathcal{GP}_{\mathrm{fail}}}\left[\mathcal{H}\left(-\mathcal{GP}_{\mathrm{fail}}\left(\mathbf{x}\right)\right)\right],
\end{equation}

where $\mathcal{H}$ is the Heaviside function. We add this to our acquisition function to get

$$
\alpha_{\mathrm{CARBS}}\left(\mathbf{x}\right)=\alpha_{\mathrm{EI-th}}\left(\mathbf{x}\right)P_{\mathrm{search}}\left(\mathbf{x}\right)P_{\mathrm{success}}\left(\mathbf{x}\right).
$$

\textbf{Cost ceiling.} We add a cost maximum so that CARBS will not suggest parameters that take too long to evaluate. We implement this with a hard threshold by removing suggestions where $\tilde{c}>c_{max}$.

\textbf{Parallelism.} We would like to have multiple different suggestions under evaluation in parallel. The above algorithm only depends on the past observations, and so it is likely to make similar suggestions until new data is obtained. To avoid this, we use Thompson sampling to sample from the GP posterior, which we then use to obtain predicted outputs $\hat{y}_{i}$ from $\mathcal{GP}_{y}$ for the outstanding suggestions $\mathbf{x}_i$ (parameters that have been suggested but not observed). We then train the output model $\mathcal{GP}_{y}$ on these predicted outputs together with the observations.

\section{Experiments}
\label{sec:experiments}

\subsection{Solving the ProcGen Benchmark with PPO}

\begin{figure}[tb]
\vskip 0.2in
\begin{center}
\centerline{\includegraphics[width=0.99\columnwidth]{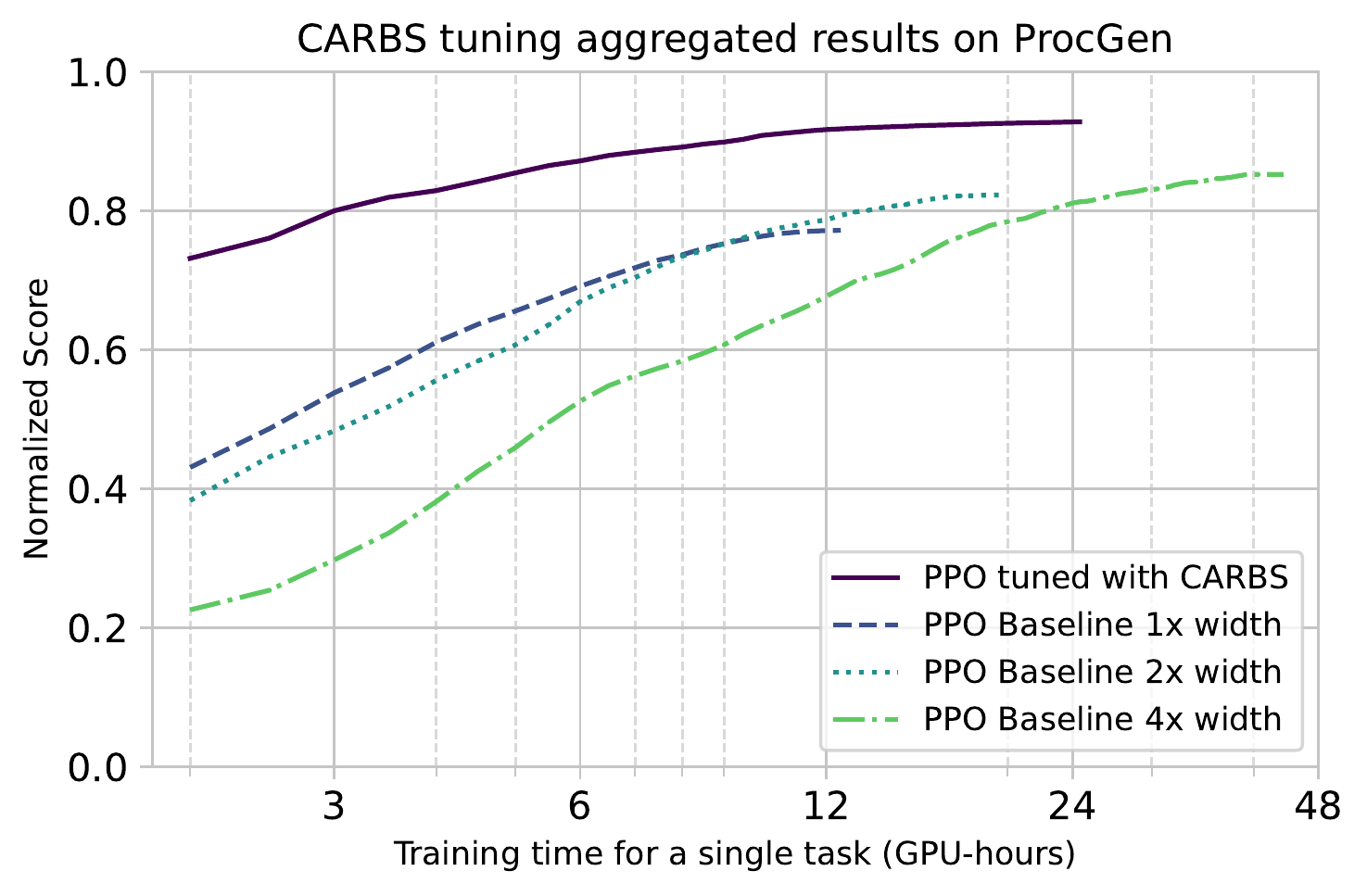}}
\caption{Aggregate normalized score vs. training time per task. The solid purple curve is the Pareto front of the CARBS-tuned PPO while the dashed lines are the learning curves of the baselines (the previous state-of-the-art performance on ProcGen). ProcGen is usually tuned for 200M steps, corresponding to 12 hours in our setup. CARBS tuning can improve performance by $>$16\% in the same amount of training time or match it in 4x less time.}
\label{fig:procgen-cost-pareto}
\end{center} 
\vskip -0.2in
\end{figure}

\begin{figure}[tb]
\vskip 0.2in
\begin{center}
\centerline{\includegraphics[width=0.99\columnwidth]{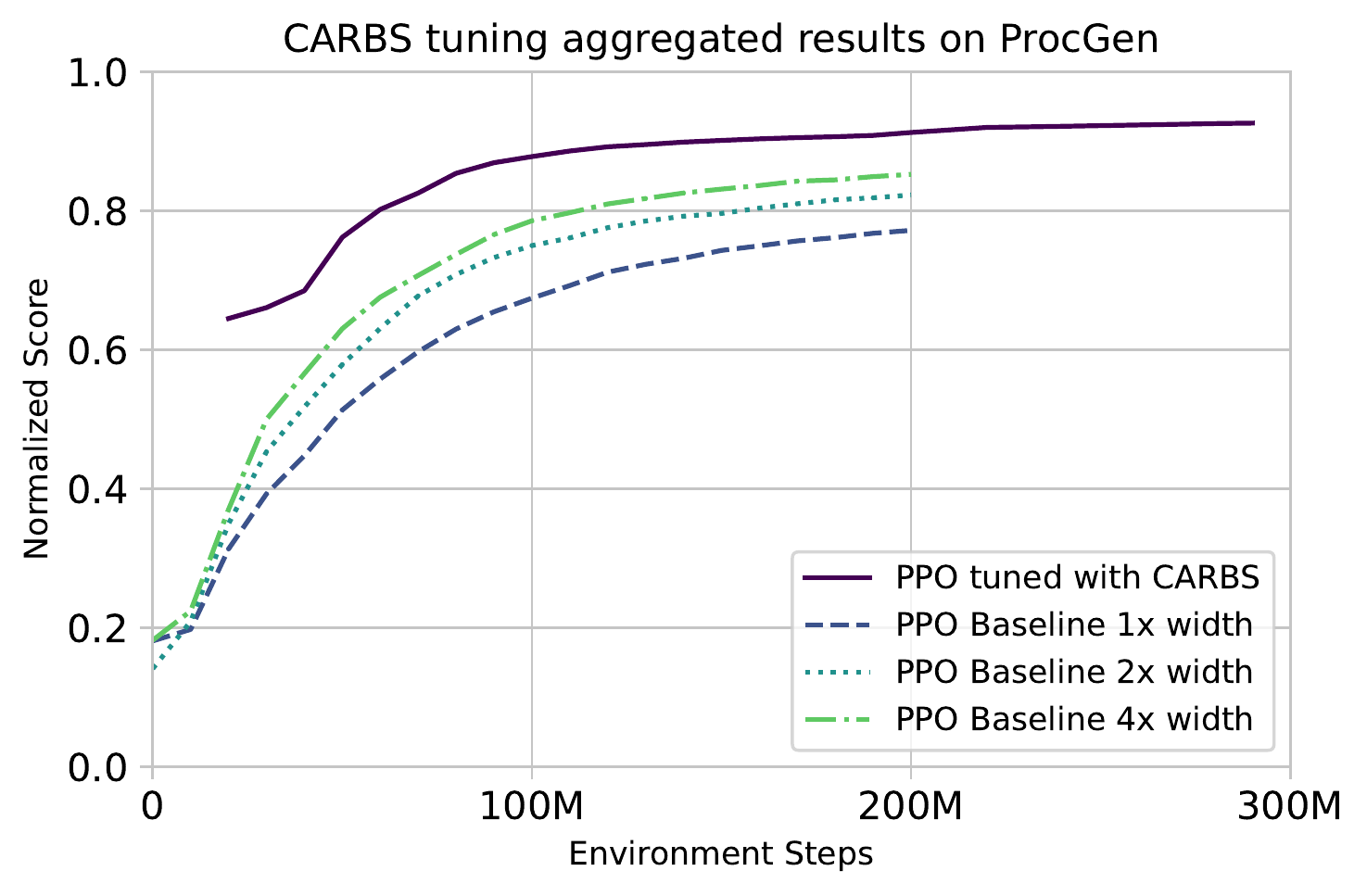}}
\caption{Aggregate normalized score vs. environment steps per task. The solid purple curve is the Pareto front of the CARBS-tuned PPO while the dashed lines are the learning curves of the baselines (the previous state-of-the-art performance on ProcGen). ProcGen is usually tuned for 200M steps. CARBS tuning can match the state-of-the-art performance with 1.5-4x higher sample efficiency.}
\label{fig:procgen-sample-pareto}
\end{center}
\vskip -0.2in
\end{figure}

Reinforcement learning algorithms are notoriously difficult to tune as they have many hyperparameters, including cost-influencing parameters such as the number of parallel environments, the number of passes over the data, and the network width. Further complicating tuning, reinforcement learning environments can be very noisy. 

To test whether CARBS can consistently improve performance in this type of environment, we tune 13 hyperparameters (see Appendix Table~\ref{tab:params-procgen}) of a PPO baseline on each of the 16 tasks of the ProcGen benchmark. We find that CARBS dramatically improves both performance and training cost on all tasks that are not already solved by the baseline, in many cases fully solving them. For the three tasks that are already solved by the baseline, CARBS still significantly reduces the time required to solve the task. To the best of our knowledge, this CARBS-tuned PPO beats all other state-of-the-art results on ProcGen.

In detail: The results from \citealt{Cobbe++19} are still the best-performing published PPO hyperparameters, so we use that as our baseline. We also include the 2x and 4x width configurations as additional baselines. We stress that these baselines were tuned by the original authors, just not with CARBS. We normalize the per-task scores by dividing by the maximum theoretical score on each task, although a perfect score is not necessarily achievable even by humans.

Figure~\ref{fig:procgen-cost-pareto} shows the Pareto front found by CARBS compared with the learning curves of the baselines. The ProcGen benchmark is typically trained for 200M environment steps. Our setup with one NVIDIA 3090 GPU and 8 CPU cores is able to train the 1x width baseline in about 12 hours per task and achieves an aggregate normalized score of 0.79. After tuning each task with CARBS with 8 parallel workers for ~9 days, the aggregate performance in 12 hours per task improves to 0.92, an increase of more than 16\%. In addition to better performance at 12 hours, CARBS finds a configuration that can get to the same 0.79 performance 4x faster. Results for individual tasks are shown in Appendix Figure~\ref{fig:procgen-tasks-cost}.

Although we did not explicitly search for algorithms with improved sample efficiency, we are able to use the observations from our tuning results to construct another Pareto curve trading-off score with number of environment steps, shown in Figure~\ref{fig:procgen-sample-pareto}. These results demonstrate that CARBS-tuned PPO can get to the same performance with 1.5-4x less data than the baseline. Results for individual tasks are shown in Appendix Figure~\ref{fig:procgen-tasks-env-steps}. 

\begin{figure*}[tb]
\vskip 0.2in
\begin{center}
\centerline{\includegraphics[width=1.0\linewidth]{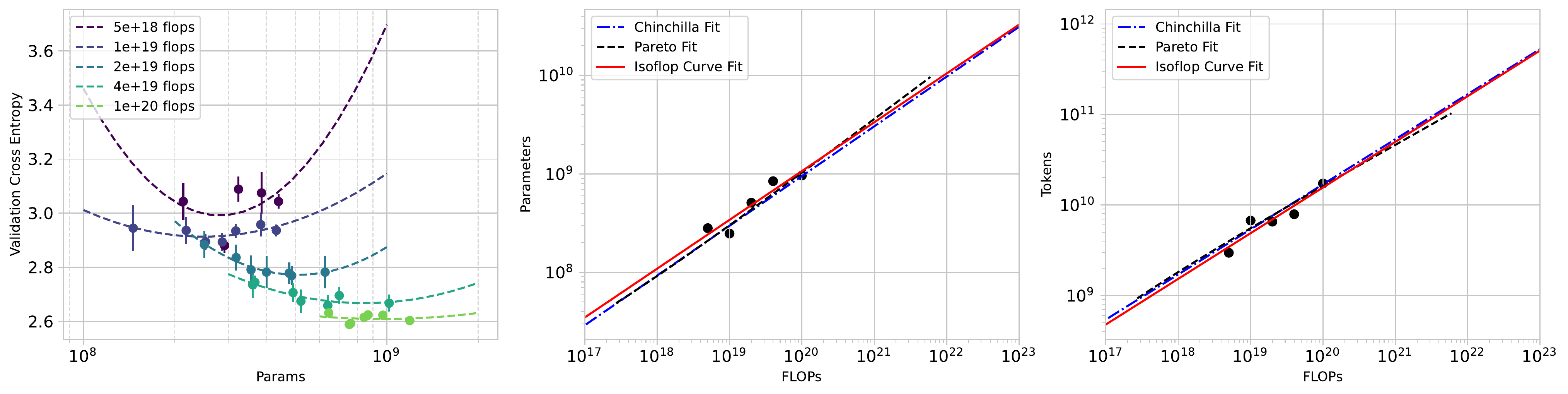}}
\caption{Left: Approximate isoflop curves for GPT-like Transformers trained on C4. Middle and Right: Scaling predictions for optimal parameters and tokens based on fit isoflop curves (blue circles and line) and a linear fit to the Pareto front (black dashed lines).}
\label{fig:isoflop-curves}
\end{center}
\vskip -0.2in
\end{figure*}

\begin{figure*}[h!]
\vskip 0.2in
\begin{center}
\centerline{\includegraphics[width=1.0\linewidth]{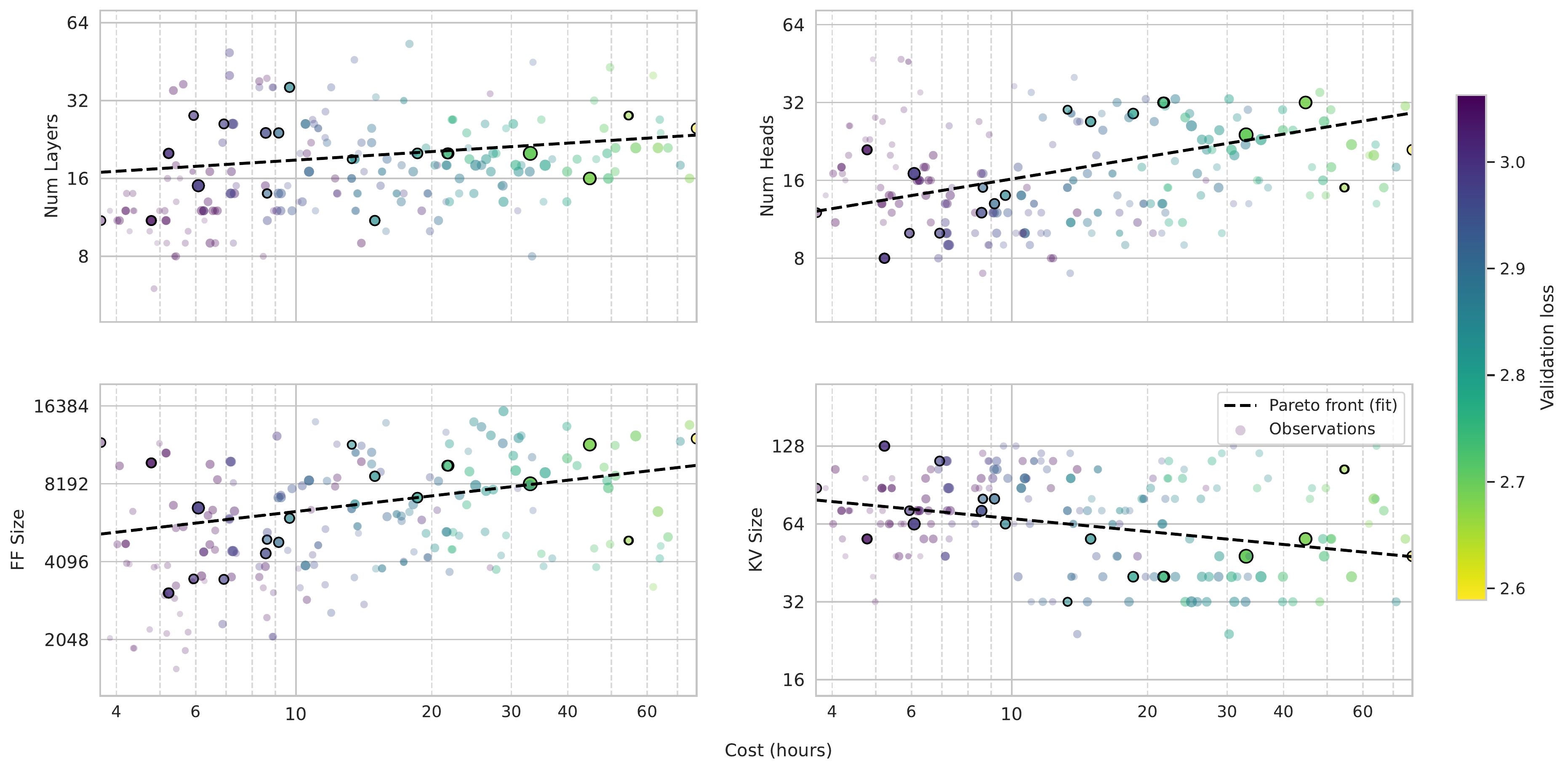}}
\caption{Fit scaling predictions for different model size parameters from a CARBS run on GPT-like Transformers trained on C4. The color of the observations reflects the final validation loss of each run on the C4 dataset. The size and transparency of each data point reflect the distance to the fit Pareto front (smaller and lighter means further away). The observations with the black border are part of the Pareto front. }
\label{fig:scaling-curves}
\end{center}
\vskip -0.2in
\end{figure*}

\subsection{Scaling laws for language modeling}

With enough data, the GP model CARBS uses can automatically discover scaling laws and leverage them to improve performance. We test the discovery of scaling laws with CARBS by tuning GPT-like Transformers to minimize validation loss on the C4 dataset, and compare our predicted scaling laws with those reported in the Chinchilla paper~\cite{Hoffmann++22}. We find that the predictions are very similar, with a detailed investigation matching the Chinchilla results exactly.

We used a Transformer implementation from Mosaic ML~\cite{mosaicml2022}, together with some implementation details to better match Chinchilla: we use the SentencePiece tokenizer \texttt{t5-small}~\cite{2020t5}, use relative positional encodings~\cite{shaw2018self}, and set the initialization to be the same as T5~\cite{2020t5}. We also used a few additional parameters for initialization we attribute to the muP scaling~\cite{Yang++22}: separate initialization scale for the embedding and Transformer, as well as scale parameters we tuned on the attention softmax and final output. Together, we tuned 19 different parameters including regularization terms, schedule, model size and token count.

We ran CARBS with 8 workers in parallel for 35 days, setting CARBS to suggest only parameters which would take less than 5 days of training time. Each worker was a single machine with 8x 40GB A100 GPUs. We started tuning at a model size of 125M parameters.

We set out to replicate \textit{Approach 2} proposed in the Chinchilla paper by taking isoflop slices from our pool of observations. Figure~\ref{fig:isoflop-curves} shows approximate isoflop curves that are derived from the data using a method described in Appendix~\ref{appendix:gpt-isoflop}. A linear fit to the minima of each of these isoflop curves yields $a=0.50$ for $N_{opt}\propto C^a$ and $b=0.50$ for $D_{opt}\propto C^b$. The Chinchilla results for $a$ and $b$ were also 0.50 (see their Table A2), indicating exact agreement with our findings despite the noisy data.

We used a simplified method to predict scaling laws for all of the parameters being searched over. We simply fit a linear regression model to all of the parameters of the search space in their transformed space, using only the Pareto front observations as input. This model is shown in Fig~\ref{fig:isoflop-curves} and ~\ref{fig:scaling-curves} as a black dashed line. One can see that it is only slightly divergent from the previous prediction in Figure~\ref{fig:isoflop-curves}, with a slope around 0.51 to 0.55 for $a$ and 0.45 to 0.49 for $b$. 

We show how this model fits different terms related to the model size in Figure~\ref{fig:scaling-curves}. Although the data is quite noisy, we can see that the best fit is to not just uniformly increase the model size. The parameter most sensitive to scale is the width of the feed-forward MLPs, with the network depth increasing more slowly. More intriguingly, the model dimension $d_{model}=d_{kv}\cdot n_{head}$ is held almost constant, with a \textit{decrease} in the size of the attention $q,k,v$ dimensions at larger scale. We include this type of plot for every search parameter in Appendix~\ref{appendix:gpt-param-scaling}.

For our analysis, CARBS observed 340 training runs, with models from 19M to 1.5B parameters trained for 600M to 24B tokens. By comparison, Chinchilla's analysis required training 50 models ranging from 44M to 16B parameters (see their Table A9) with 9 different learning rate schedules/number of training tokens and held all other hyperparameters fixed. We thus conclude that the CARBS analysis required less compute despite also marginalizing over many other hyperparameters and obtaining scaling relationships for each of them as well.

\subsection{Comparison with other tuners}

CARBS provides a means to thoroughly and robustly tune large, complex, noisy models with non-trivial hyperparameter search spaces, of the sort that are historically difficult to tune yet are of increasing research interest. This makes comparisons to existing tuners challenging. Nevertheless, for completeness, we provide a comparison to four existing algorithms on smaller problems that these methods are better suited to handle. We emphasize that the main goal of CARBS is not to move the needle on these types of problems but rather to enable tuning on problems that are intractable with current methods. 



We design three smaller tuning problems across different domains: language modeling, reinforcement learning, and image classification. We compare against four tuners, chosen for their popularity, existing implementations in the Ray Tune library~\cite{liaw2018tune}, and diversity. The details of the tasks and baselines are given below.

\subsubsection{Tasks}

\textbf{Language modeling.} We tokenize the \texttt{WikiText103} dataset~\cite{merity++16} using whole words, and train a masked Transformer language model~\cite{vaswani2017attention} according to a PyTorch example script~\cite{pytorch+Transformer}. We use defaults primarily from the \texttt{gpt-micro} setting of \texttt{mingpt} as our initial search center~\cite{mingpt}.

\textbf{Reinforcement learning}. We train a PPO \citealt{Schulman++17}) policy on the \texttt{Ant-v4} MuJoCo gym environment~\cite{Brockman++16} using envpool for parallelism~\cite{envpool}. We use a recent implementation of PPO that has carefully reproduced published results~\cite{shengyi++2022}. We use default parameters from that script as our initial search center.

\textbf{Image classification.} We train convolutional neural network models to classify images from the imagenette dataset~\cite{imagenette}. We use the PyTorch example script~\cite{pytorch+imagenet} for training, with the default torchvision ResNet~\cite{he2016deep} modified to allow us to vary the depth and width. We also add random augmentations~\cite{cubuk2020randaugment} from \texttt{torchvision} and label smoothing~\cite{szegedy2016rethinking}. We use default parameters from the original script and from the ResNet-RS paper as our initial search center~\cite{ResNet-RS}.

We explicitly include cost-related parameters like the model size, batch size, and the amount of training data in the search space of around 10 parameters for each problem. All parameters we tuned are described in Appendix~\ref{appendix:comparison}. The tuning experiments were run with 8 independent workers with a single GPU each, and the best set of parameters was chosen after 12 hours of tuning time. This is a significantly higher compute requirement than other baselines like Bayesmark~\cite{Turner++21}, which is more representative of practical problems in machine learning.


\subsubsection{Baselines}

\textbf{Blended search,} an algorithm that utilizes local search starting at a set of specified low cost parameters~\cite{Wang++21}. It combines local search with a global algorithm for starting and stopping the local search threads.

\textbf{HEBO}, the winner of the 2020 NeurIPS Black Box Optimization Challenge~\cite{Cowen-Rivers++20, Turner++21}. It utilizes a GP model of past observations as well as leveraging several BO acquisition functions in parallel for suggestion.

\textbf{Tree of Parzen Estimators (TPE),} a popular algorithm for hyperparameter tuning that maintains a tree-based model of all past observations~\cite{Akiba++19}. 

\textbf{Asynchronous Successive Halving Algorithm (ASHA),} a variant of random search that terminates the lowest performing runs early~\cite{Li++18}. By doing so, it is able to sample many more parameters than other algorithms.

\subsubsection{Fairness of comparison}

While CARBS only requires specifying an initial search center for each parameter, the baseline algorithms require finite search bounds. We choose these bounds by running CARBS three times on each problem and choosing minimum and maximum values for each search parameter that included the observations that ran successfully (see Appendix \ref{appendix:comparison}). 

We endeavor to make the comparison as fair as possible, avoiding choices that give CARBS an advantage (e.g. by using wider default search bounds instead of the search subspace that CARBS discovers, or by adding more hyperparameters to tune). Additionally, these baseline algorithms have no way of modeling training failures, such as out-of-memory errors, so we return a constant result for failed runs corresponding to the worst observed result for that task.




\subsubsection{Results}
\label{subsec:exp-comparison}

\begin{figure}[t!]
\vskip 0.2in
\begin{center}
\centerline{\includegraphics[width=\columnwidth]{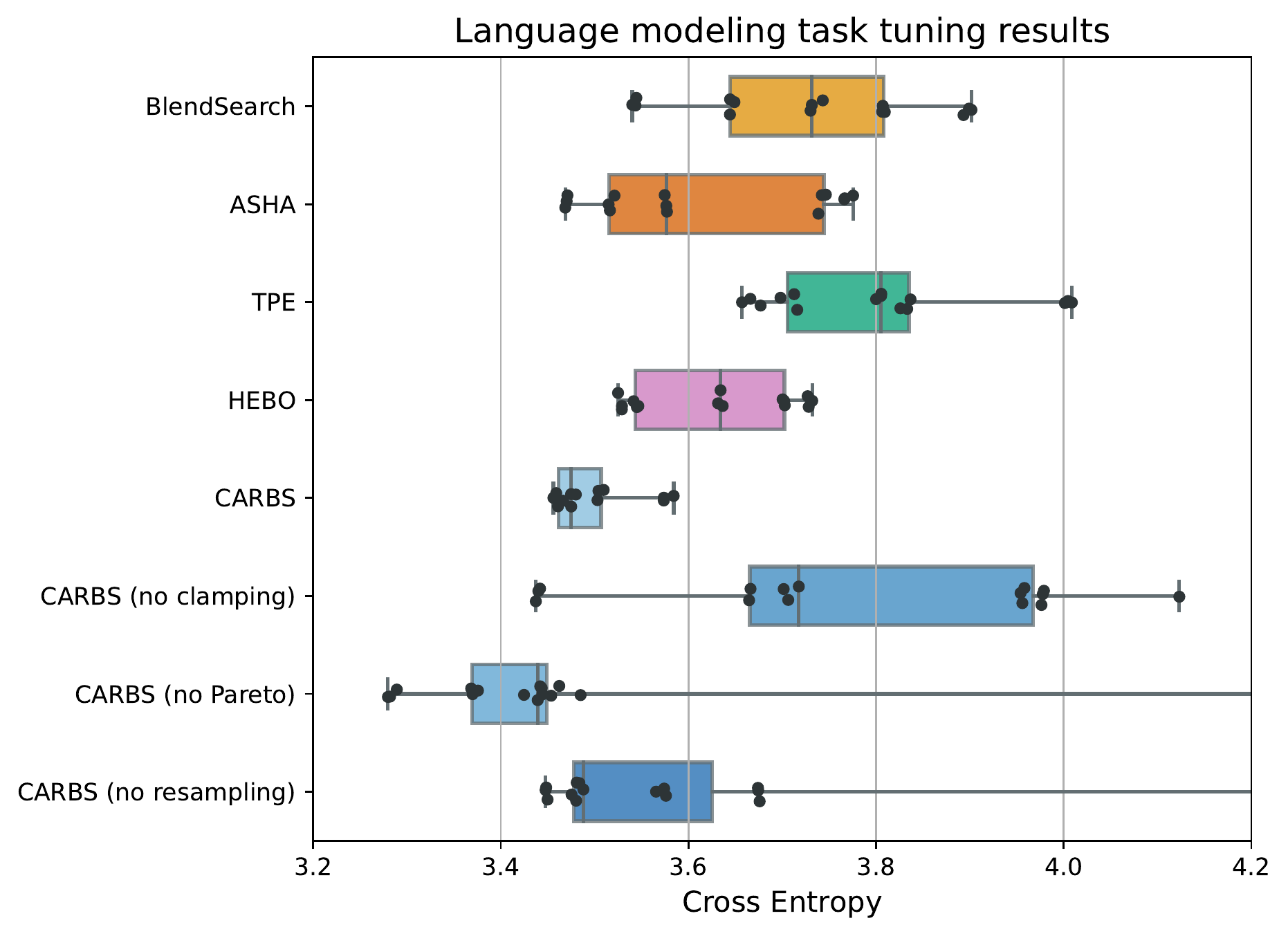}}
\vskip 0.2in
\centerline{\includegraphics[width=\columnwidth]{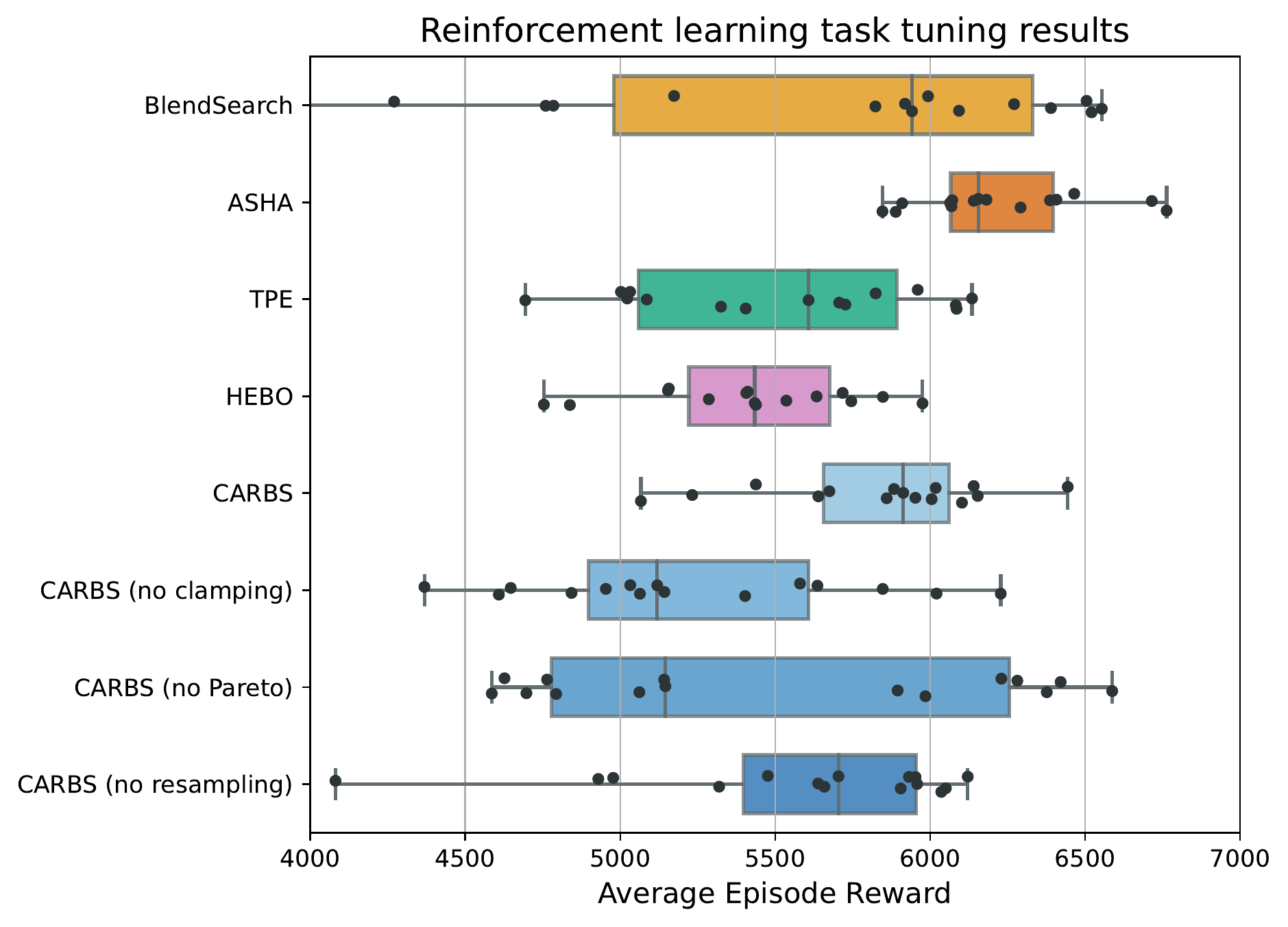}}
\vskip 0.2in
\centerline{\includegraphics[width=\columnwidth]{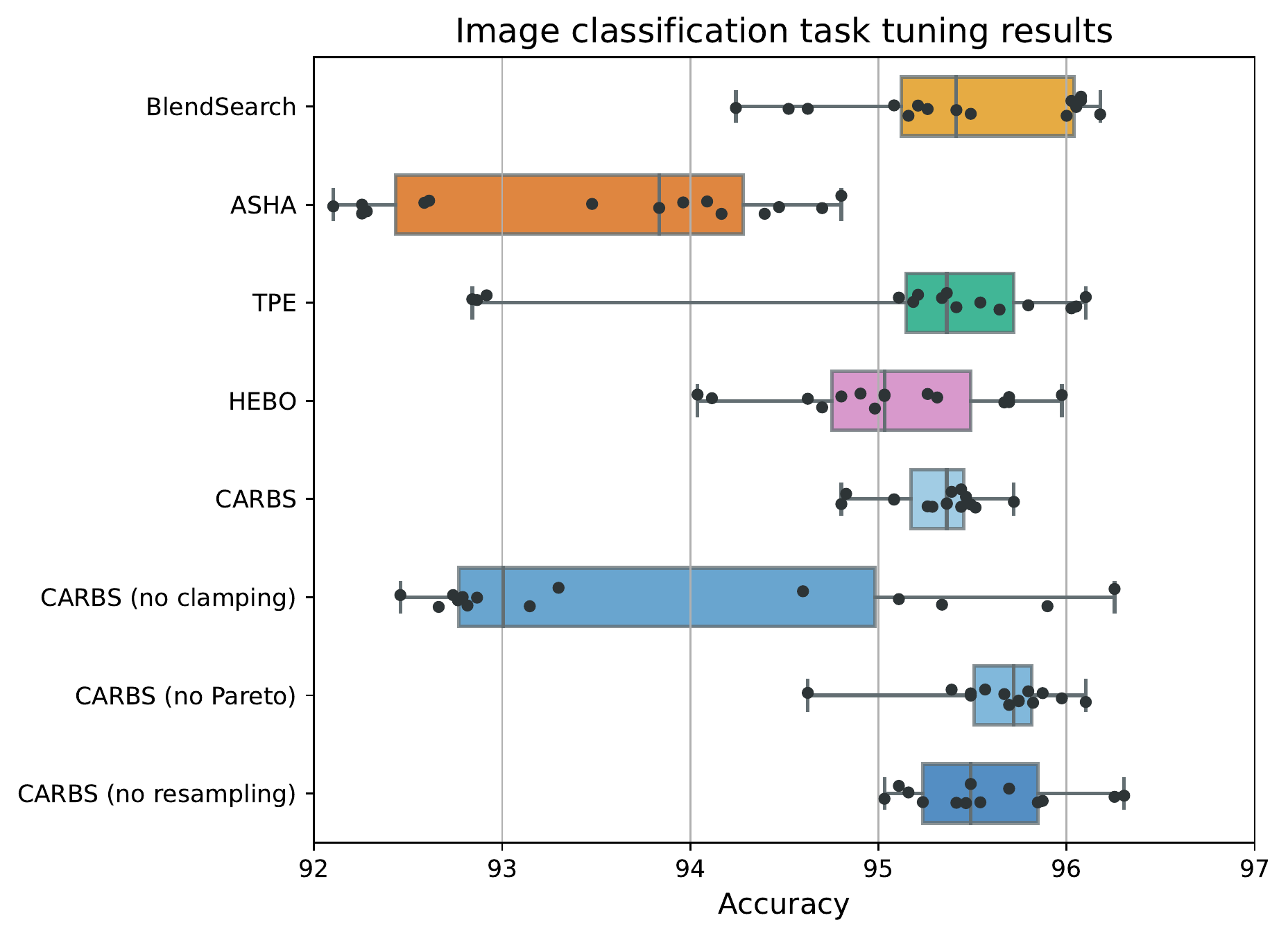}}
\caption{Final performance of different hyperparameter tuning algorithms on the language modeling task (top), reinforcement learning task (middle) and image classification task (bottom). Each algorithm was run five times and the best set of parameters was rerun three times.}
\label{fig:boxplots}
\end{center}
\vskip -0.2in
\end{figure} 

We find that CARBS performs comparably to the other algorithms on all tasks (see Figure~\ref{fig:boxplots}), and is the best-scoring algorithm on the language modeling task. 

In addition to high median performance, CARBS produces a much lower variance distribution of outputs. This robust tuning is particularly useful when comparing two tuned algorithms with each other. Similarly, we note that CARBS performance is consistent across tasks whereas e.g. ASHA is the highest performing algorithm on the reinforcement learning task but the lowest on the image classification task.

\begin{figure*}[tb]
\vskip 0.2in
\begin{center}
\centerline{\includegraphics[width=1\linewidth]{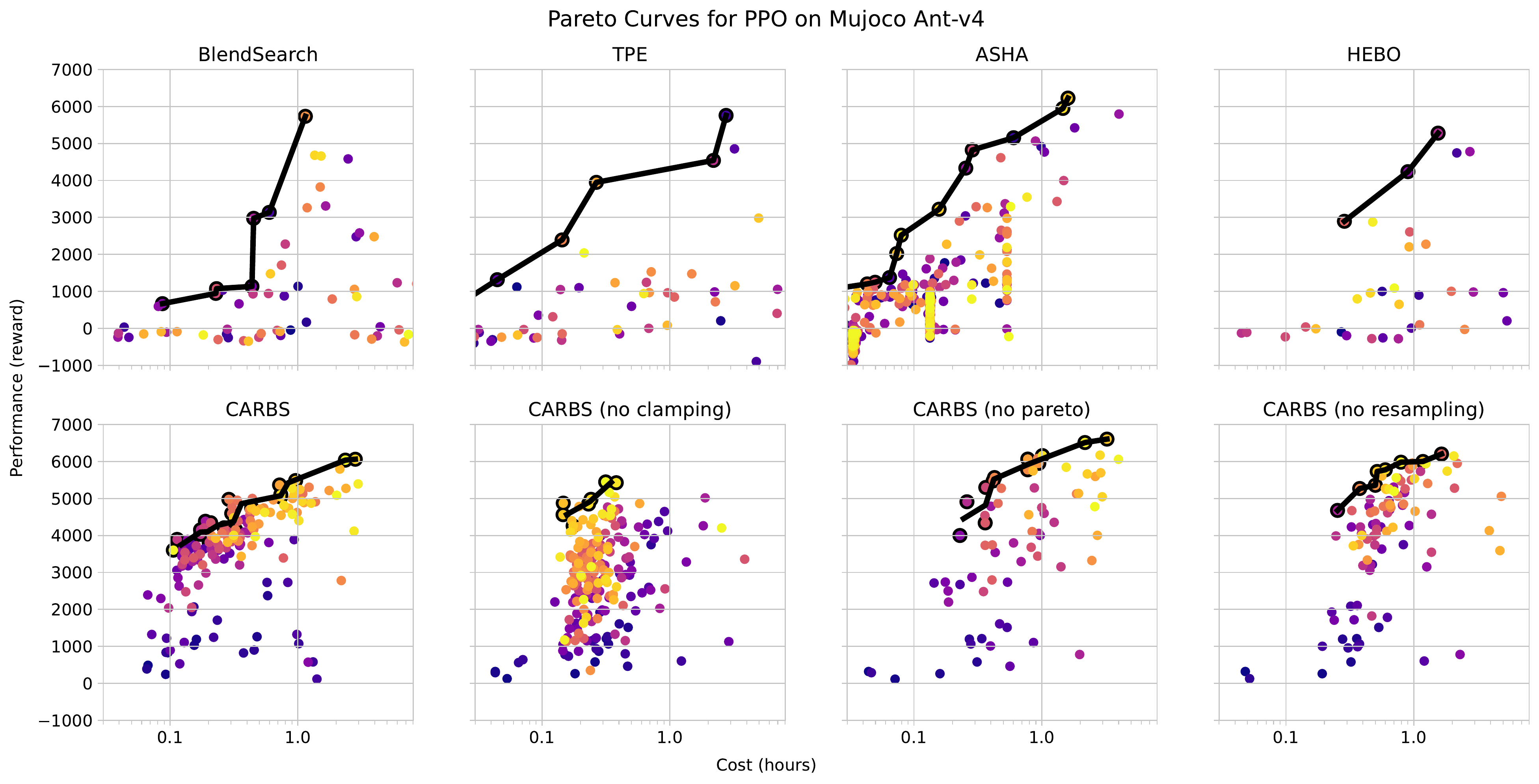}}
\caption{Observations (dots) and Pareto fronts (black lines) for a single 12 hour, 8-worker training run tuning PPO on MuJoCo \texttt{Ant-v4} for a variety of training algorithms. The color of the observations indicates the order observations were made, from purple at the start of training through magenta, orange, and yellow for the final observations. CARBS has more total observations because it uses its budget more efficiently.}
\label{fig:ppo-paretos}
\end{center}
\vskip -0.2in
\end{figure*}

With the search spaces we have defined, it is useful to compare tuning runs not just in terms of best overall performance, but also in terms of how long it takes to achieve that performance since training time can vary by several orders of magnitude. We compare the observations and the resulting Pareto fronts for different tuning algorithms on the reinforcement learning task in Fig~\ref{fig:ppo-paretos}. 

Compared to other hyperparameter tuning algorithms, CARBS produces many more samples close to the Pareto front. This is useful for producing a better surrogate model and for making predictions for how to best scale parameters. While observations far from the Pareto front narrow down the global hyperparameter space, they do not contribute to a useful model of performance near the Pareto front.

We also note the interesting failure modes of different CARBS ablations depicted in these charts. The version of CARBS that always uses the maximum observation for the EI (no Pareto) does not sample as many points close to the Pareto front, as it samples more high cost points. Although this does sometimes lead to higher performance, it is less robust than the other acquisition function, as can be seen in the aggregated box plots (Fig~\ref{fig:boxplots}). The version that uses the acquisition function $\alpha_\mathrm{EI-pf}$ without the random cost clamping (no clamping) includes more low cost points at the expense of exploring higher cost, higher performing ones. And the version of CARBS without resampling (no resampling) does not discover the same cost-performance scaling relationship because of the noisy observations.

\section{Conclusion}
\label{sec:conclusion}

We presented CARBS, a new Bayesian optimization algorithm for hyperparameter tuning that models the performance-cost Pareto frontier. CARBS benefits from a local search strategy in the vicinity of the current Pareto set at each iteration, which enables it to efficiently optimize even when the number of hyperparameters is large, and also benefits from its cost-awareness, which helps it automatically learn how to tune as it scales. We demonstrated that simple algorithms tuned with CARBS achieve breakthrough performance, and that scaling laws can be derived from the tuning process without additional effort. Though rigorous manual scaling studies are the best way to have confidence in extrapolating to larger scales, we hope CARBS will provide an easier way to do this on a lot of problems.

\bibliography{references}
\bibliographystyle{icml2023}

\clearpage

\newpage
\appendix
\onecolumn


\section{CARBS Algorithm}
\label{appendix:algorithm}

\begin{algorithm}
   \caption{Simplified CARBS Algorithm}
   \label{alg:carbs-simple}
\begin{algorithmic}
   \STATE {\bfseries Input:} search center $\mu$, search radius $\sigma_{\rm search}$, number of candidates per pareto front point $N_{\rm cand}$
   \STATE Add observations to $X_i$
   \REPEAT
   \STATE Sample suggestions from $\mathcal{N}(\mu,\sigma_{\rm search})$.
   \UNTIL{$X_i$ has observations}
   \REPEAT
   \STATE Calculate the pareto front $PF$ from $X_i$ using Eq.~(\ref{eq:pf})
   \STATE Sample $N_{\rm cand}$ candidate points $X_{\rm cand}$ from Gaussian distributions of radius $\sigma_{\rm search}$ around each point in $PF$
   \STATE Fit Gaussian process models $\mathcal{GP}_y$, $\mathcal{GP}_c$, and $\mathcal{GP}_\mathrm{pf}$ with observations $X_i$ and $PF$  according to Eq.~(\ref{eq:gps})
   \FOR{each candidate point}
   \STATE Calculate expected costs $\tilde{c}$ , outputs $\tilde{y}$, and pareto front values $\tilde{y}_\mathrm{pf}$
   \STATE Evaluate the acquistion function $\alpha_{\mathrm{EI-pf}}$ using Eq.~(\ref{eq:ei-pf})
   \ENDFOR
   \STATE Return candidate point with highest acquisition function value
   \UNTIL{terminated}
\end{algorithmic}
\end{algorithm}

\section{ProcGen Experiment}
\label{appendix:procgen}

\subsection{Search parameters}

\begin{table}[h]                           
 \centering
\begin{tabular}{llrrr}
\hline
 name            & Space type   &   Search center &   min & max      \\
\hline
 total env steps & Log          &          2e+07  & 1e+06 & $\infty$ \\
 num steps       & Log          &        256      & 1     & 1024     \\
 num workers     & Log          &        256      & 1     & 1024     \\
 minibatch size  & Log          &       8192      & 1     & 8192     \\
 ppo epochs      & Linear       &          3      & 1     & $\infty$ \\
 lr              & Log          &          0.0005 & 0     & $\infty$ \\
 clip range      & Logit        &          0.2    & 0     & 1        \\
 lam             & Logit        &          0.95   & 0     & 1        \\
 discount        & Logit        &          0.999  & 0     & 1        \\
 entropy coef    & Log          &          0.01   & 0     & $\infty$ \\
 value loss coef & Log          &          0.5    & 0     & $\infty$ \\
 cnn base width  & Log          &         16      & 1     & $\infty$ \\
 mlp width       & Log          &        256      & 1     & $\infty$ \\
\hline
\end{tabular}
\caption{Search parameters for tuning PPO on ProcGen.}
\label{tab:params-procgen}      
\end{table}

\subsection{Per-task scores}

\begin{figure}[h!]
\vskip 0.2in
\begin{center}
\centerline{\includegraphics[width=\columnwidth]{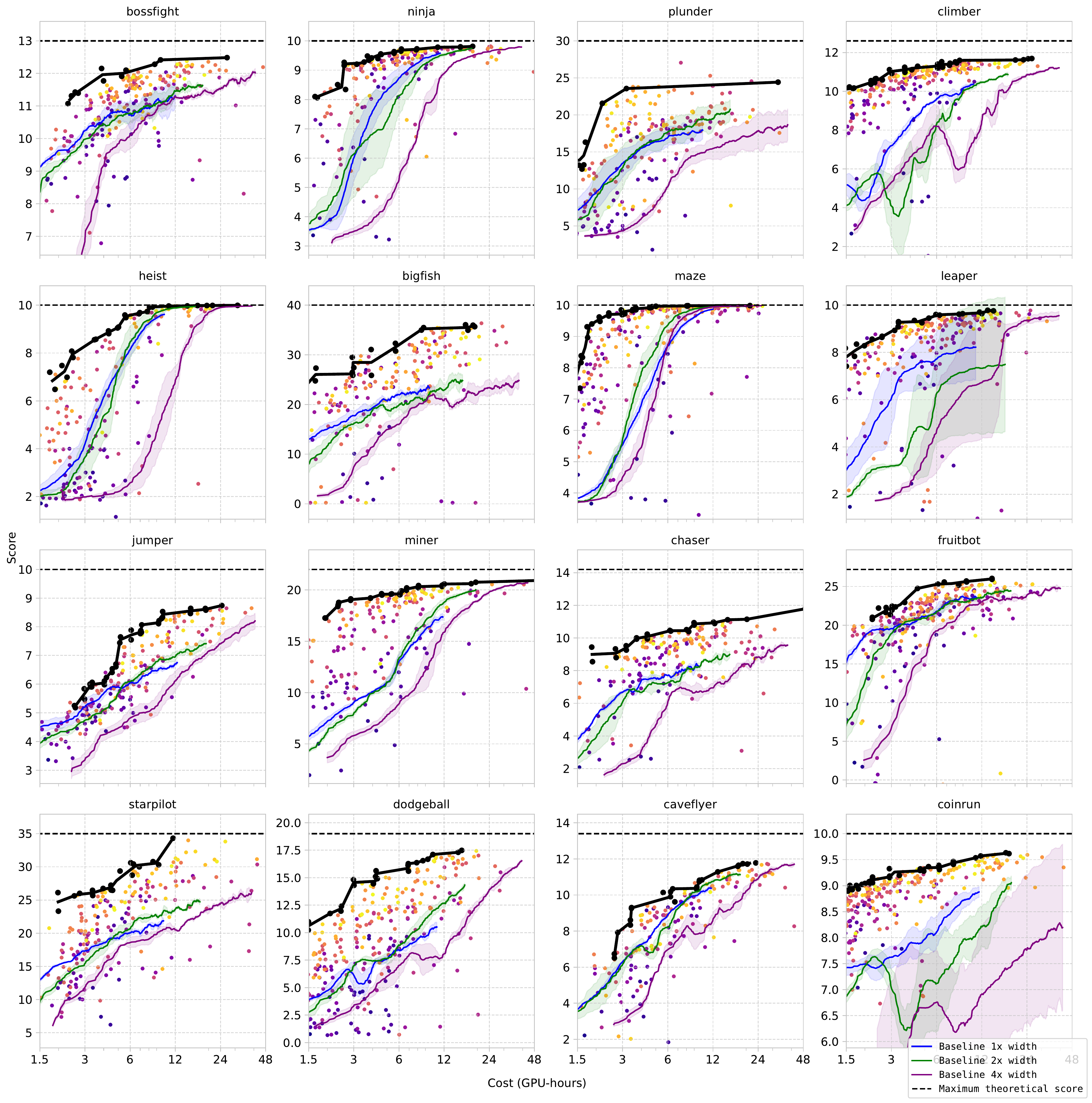}}
\caption{Performance-cost Pareto curves on each ProcGen task after tuning PPO with CARBS.}
\label{fig:procgen-tasks-cost}
\end{center}
\vskip -0.2in
\end{figure} 

\begin{figure}[h!]
\vskip 0.2in
\begin{center}
\centerline{\includegraphics[width=\columnwidth]{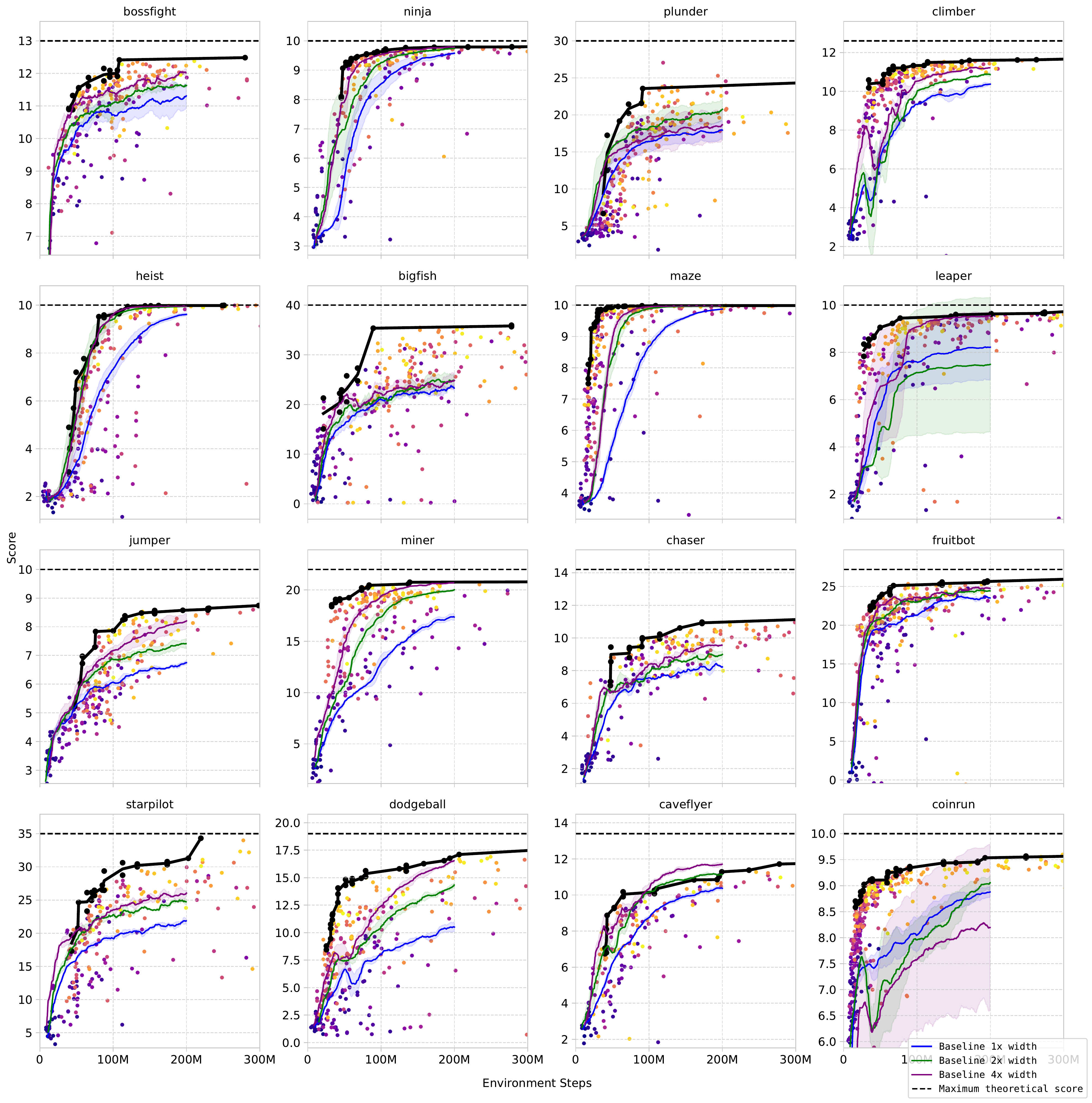}}
\caption{Performance-sample count Pareto curves on each ProcGen task after tuning PPO with CARBS.}
\label{fig:procgen-tasks-env-steps}
\end{center}
\vskip -0.2in
\end{figure} 

\clearpage

\section{GPT Tuning and Scaling Laws}
\subsection{Isoflop curve method}
\label{appendix:gpt-isoflop}

We set out to replicate the results of the Chinchilla paper by taking isoflop slices from our pool of observations. This is a bit tricky, as the observations we made vary along 19 different dimensions. We approximate the number of flops as 6ND, where N is the number of model parameters and D is the number of tokens

To get each isoflop curve, we first select observations near the target flop count (from 0.8x to 1.2x). We fix most of those input parameters, and only adjust the token count such to match the target flop count. We then pass those flop-adjusted parameters to our GP model to get a target-flop corrected output, as well as error bounds from the model. Finally, to reduce the effect of models that are poorly tuned in other dimensions, we split the observations into $k$ bins by their parameter count, and choose the best output for each bin. We fit each curve in the manner of Chinchilla, and plot each of these isoflop curves in Figure~\ref{fig:isoflop-curves}.

\subsection{Scaling for all parameters}
\label{appendix:gpt-param-scaling}

\begin{table}[h]                           
 \centering
\begin{tabular}{llrrr}
\hline
 name               & Space type   &   Search center &   min & max      \\
\hline
 learning rate      & Log          &         0.0004  & 0     & $\infty$ \\
 num train tokens   & Log          &         2.5e+09 & 1e+08 & $\infty$ \\
 num layers         & Log          &        12       & 2     & $\infty$ \\
 num heads          & Log          &        12       & 1     & $\infty$ \\
 kv size            & Log          &        64       & 8     & 128      \\
 ffw size           & Log          &      3072       & 8     & $\infty$ \\
 microbatch size    & Log          &        12       & 1     & $\infty$ \\
 max seq len        & Log          &      1024       & 2     & $\infty$ \\
 init std           & Log          &         1       & 0     & $\infty$ \\
 embedding init std & Log          &         1       & 0     & $\infty$ \\
 output scaling     & Log          &         1       & 0     & $\infty$ \\
 attn pdrop         & Logit        &         0.01    & 0.001 & 0.5      \\
 resid pdrop        & Logit        &         0.01    & 0.001 & 0.5      \\
 emb pdrop          & Logit        &         0.01    & 0.001 & 0.5      \\
 weight decay       & Log          &         0.1     & 0     & $\infty$ \\
 grad clip norm     & Log          &         1       & 0     & $\infty$ \\
 alpha f            & Logit        &         0.1     & 0     & 1        \\
 warmup frac        & Logit        &         0.02    & 0     & 1        \\
 attn softmax scale & Log          &         8       & 0     & $\infty$ \\
\hline
\end{tabular}
\caption{Search parameters for tuning GPT on the C4 dataset.}
\label{tab:params-gpt}      
\end{table}

\begin{figure}
\vskip 0.2in
\begin{center}
\centerline{\includegraphics[width=0.8\columnwidth]{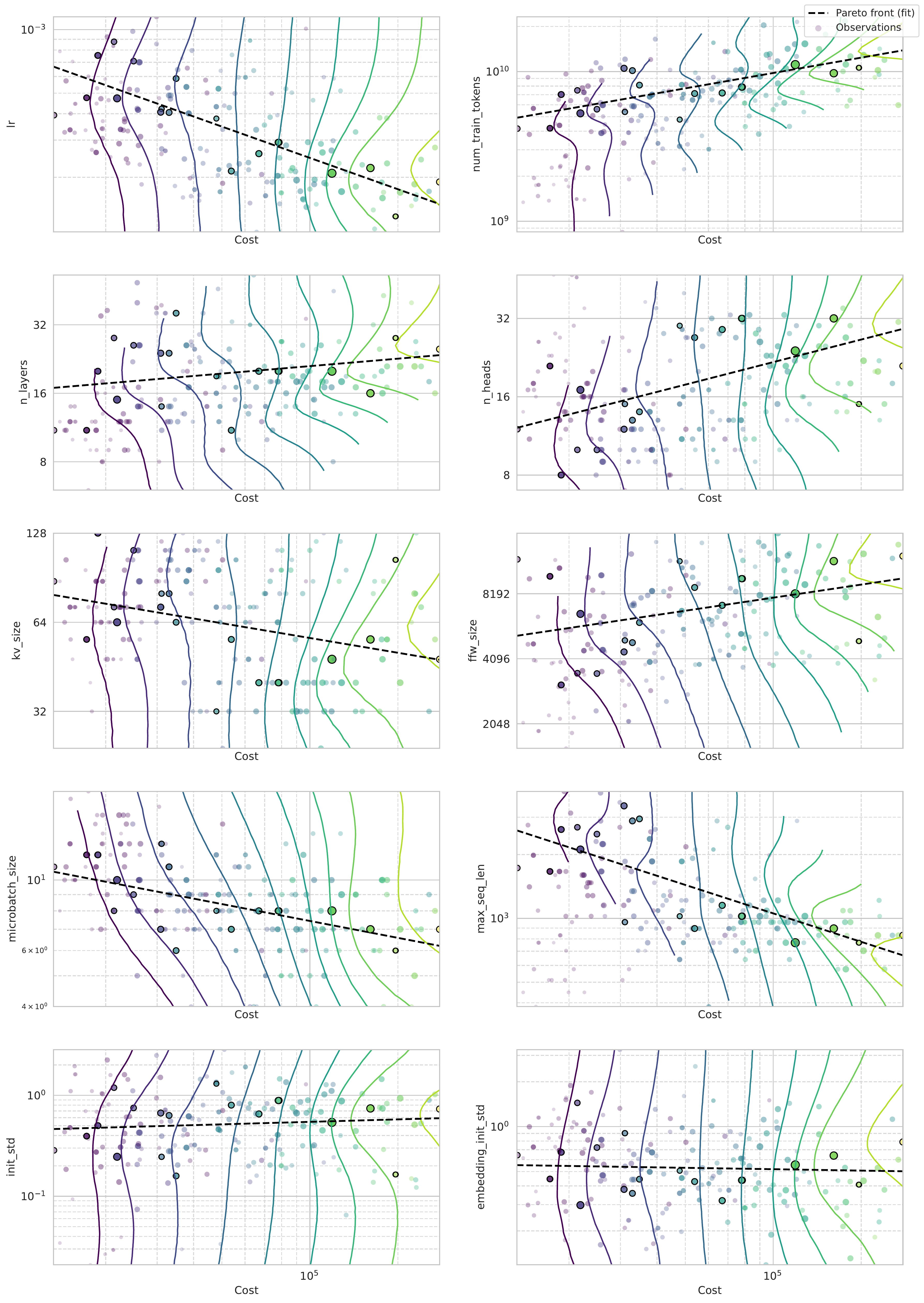}}
\caption{Fit pareto front for first half of GPT parameters tuned on the C4 dataset with CARBS. Colored lines indicate loss contours of the GP model. See other image for colorbar.}
\label{fig:gpt-params-1}
\end{center}
\vskip -0.2in
\end{figure} 

\begin{figure}
\vskip 0.2in
\begin{center}
\centerline{\includegraphics[width=0.8\columnwidth]{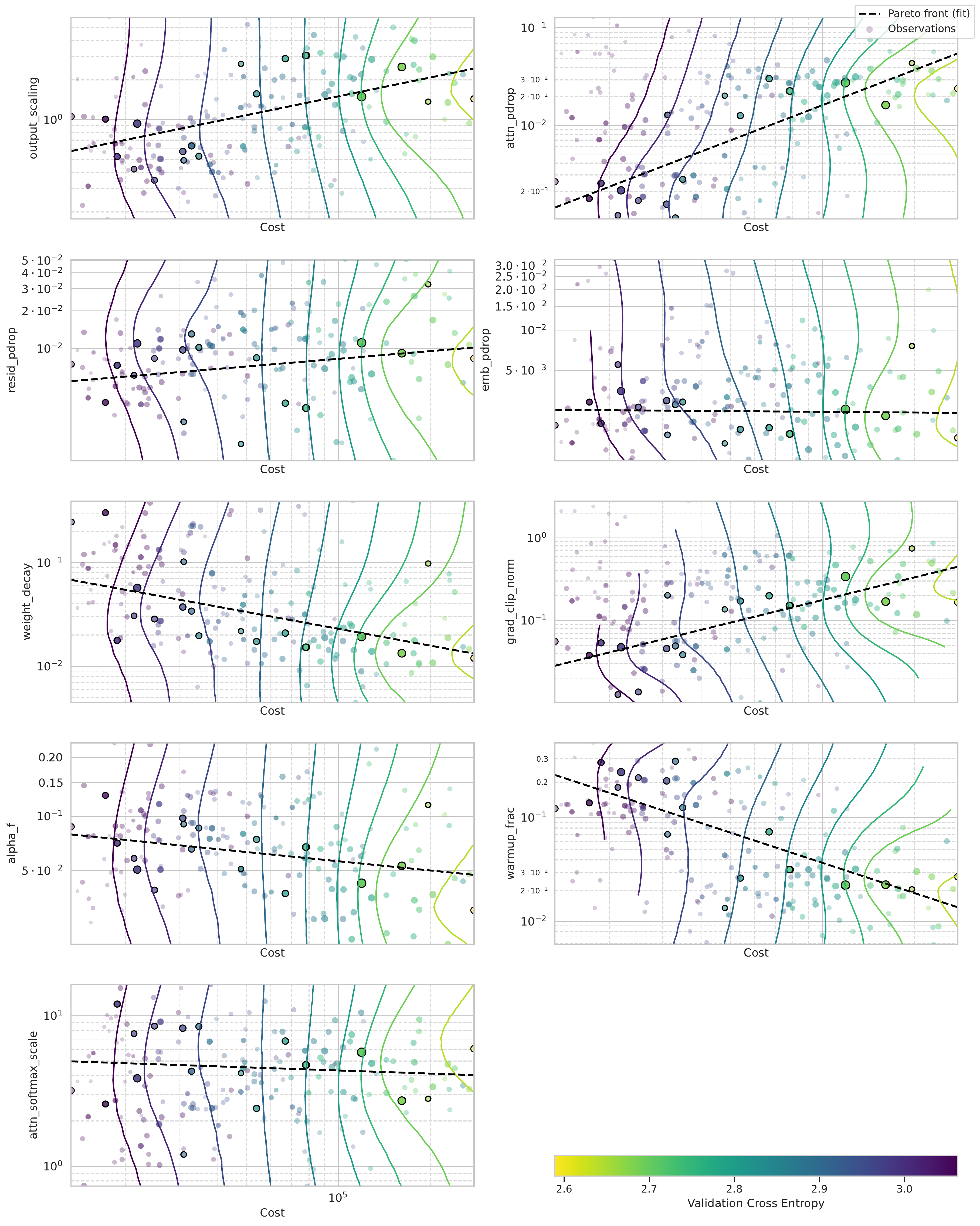}}
\caption{Fit pareto front for second half of GPT parameters tuned on the C4 dataset with CARBS. Colored lines indicate loss contours of the GP model.}
\label{fig:gpt-params-2}
\end{center}
\vskip -0.2in
\end{figure} 

\clearpage

\section{Search Parameters For Comparison Experiments}
\label{appendix:comparison}

\begin{table}[h]                           
 \centering
\begin{tabular}{llccc}
\hline
 name                   & Space type   &   Search center & CARBS min - max   & Other min - max   \\
\hline
 total timesteps & Log          &          2e+06  & 2e+05 - $\infty$  & 5e+05 - 2e+08    \\
 learning rate   & Log          &          0.0003 & 0 - $\infty$      & 1e-05 - 0.01     \\
 update epochs   & Log          &         10      & 1 - $\infty$      & 2 - 32           \\
 max grad norm   & Log          &          0.5    & 0 - $\infty$      & 0.01 - 5         \\
 num steps       & Log          &         32      & 4 - $\infty$      & 8 - 2048         \\
 num envs        & Log          &         64      & 4 - $\infty$      & 16 - 2048        \\
 num minibatches & Log          &         32      & 1 - $\infty$      & 2 - 128          \\
 clip coef       & Logit        &          0.2    & 0 - 1             & 0.01 - 0.9       \\
 gae lambda      & Logit        &          0.95   & 0 - 1             & 0.5 - 0.999      \\
 ent coef        & Log          &          0.001  & 0 - $\infty$      & 1e-05 - 0.1      \\
 gamma           & Logit        &          0.99   & 0 - 1             & 0.9 - 0.999      \\
\hline
\end{tabular}
\caption{Search parameters for reinforcement learning task. ``Other'' search bounds are used for algorithms other than CARBS.}
\label{tab:params-ppo}      
\end{table}

\begin{table}[h]                           
 \centering
\begin{tabular}{llccc}
\hline
 name                   & Space type   &   Search center & CARBS min - max   & Other min - max   \\
\hline
 kv dim         & Log          &         32      & 2 - $\infty$      & 8 - 128          \\
 nhead          & Log          &          4      & 1 - $\infty$      & 2 - 32           \\
 nhid           & Log          &        512      & 4 - $\infty$      & 256 - 5120       \\
 nlayers        & Log          &          4      & 1 - $\infty$      & 1 - 10           \\
 dropout        & Logit        &          0.1    & 0 - 1             & 5e-03 - 0.5      \\
 lr             & Log          &          0.0003 & 0 - $\infty$      & 1e-04 - 0.01     \\
 clip grad norm & Log          &          1      & 0 - $\infty$      & 0.01 - 10        \\
 train tokens   & Log          &          1e+08  & 1e+06 - $\infty$  & 1e+07 - 1e+09    \\
 batch size     & Log          &         64      & 2 - $\infty$      & 16 - 256         \\
 bptt           & Log          &        128      & 2 - $\infty$      & 32 - 1024        \\
\hline
\end{tabular}
\caption{Search parameters for language modeling task. ``Other'' search bounds are used for algorithms other than CARBS.}
\label{tab:params-ppo}      
\end{table}

\begin{table}[h!]                           
 \centering
\begin{tabular}{llccc}
\hline
 name                   & Space type   &   Search center & CARBS min - max   & Other min - max   \\
\hline
 epochs                 & Log          &         45      & 5 - $\infty$      & 12 - 1000        \\
 lr                     & Log          &          0.1    & 0 - $\infty$      & 0.01 - 1         \\
 momentum               & Logit        &          0.9    & 0 - 1             & 0.5 - 0.99       \\
 weight decay           & Log          &          0.0001 & 0 - $\infty$      & 1e-06 - 0.01     \\
 label smoothing        & Logit        &          0.1    & 1e-03 - 1         & 1e-03 - 0.5      \\
 batch size             & Log          &        256      & 16 - $\infty$     & 32 - 512         \\
 image dim              & Log          &        160      & 64 - $\infty$     & 64 - 400         \\
 num blocks             & Log          &         16      & 8 - 139           & 8 - 64           \\
 width per group        & Log          &         64      & 8 - 512           & 16 - 200         \\
 rand augment magnitude & Log          &         10      & 2 - 24            & 2 - 24           \\
\hline
\end{tabular}
\caption{Search parameters for image classification task. ``Other'' search bounds are used for algorithms other than CARBS.}
\label{tab:params-image}      
\end{table}

\end{document}